\documentclass[11pt]{article}

\usepackage[final]{acl}
\setcitestyle{authoryear,round}
\usepackage{times}
\usepackage{latexsym}
\usepackage{float}

\usepackage[T1]{fontenc}

\usepackage[utf8]{inputenc}

\usepackage{microtype}

\usepackage{inconsolata}
\usepackage{amsmath}
\usepackage{multirow}
\usepackage{cuted}
\usepackage{flushend}   
\usepackage{eso-pic}
\usepackage{graphicx}
\title{Polite but Misaligned: Evaluating LLM Politeness Judgments Against Human Pragmatic Norms}

  \author{
  Rong Wang \\
    University of T\"ubingen \\
    \texttt{rong.wang@uni-tuebingen.de}
    \And
     Kun Sun\thanks{Corresponding author.} \and Yadong Guo \\
    Tongji University \\
    \texttt{\{kunsun, guoyadong127\}@tongji.edu.cn}
     \\}

\begin{document}

\maketitle

\AddToShipoutPictureFG*{%
  \AtPageLowerLeft{%
    \raisebox{0.9cm}{%
      \makebox[\paperwidth][c]{\small\sffamily Proceedings of the 2026 Conference on Empirical Methods in Natural Language Processing (EMNLP 2026)}%
    }%
  }%
}

\begin{abstract}

Despite strong performance on standard benchmarks, it remains unclear
whether large language models (LLMs) evaluate social pragmatics in ways
that align with human judgments. We evaluate LLM politeness judgments
using two English-language datasets with complementary annotation formats: continuous human ratings and three-way categorical labels.
Across the seven evaluated models, we find that inter-model agreement is
stronger than model--human agreement. Strategy-level analyses suggest
that model--human alignment is associated with explicit linguistic cues,
while some rapport-building strategies occur more frequently in
misaligned cases. In the categorical task, model predictions exhibit
systematic neutral compression, characterized by the overproduction of
Neutral labels and the underprediction of Impolite labels. This pattern
persists when expert consensus is used as the reference on a diagnostic
subset. Our findings highlight the need for pragmatic evaluations that go beyond aggregate agreement metrics by examining directional patterns of
model--human disagreement across different human references.

\end{abstract}

\section{Introduction}
\begin{figure*}[tp]
\centering
\includegraphics[width=0.8\textwidth]{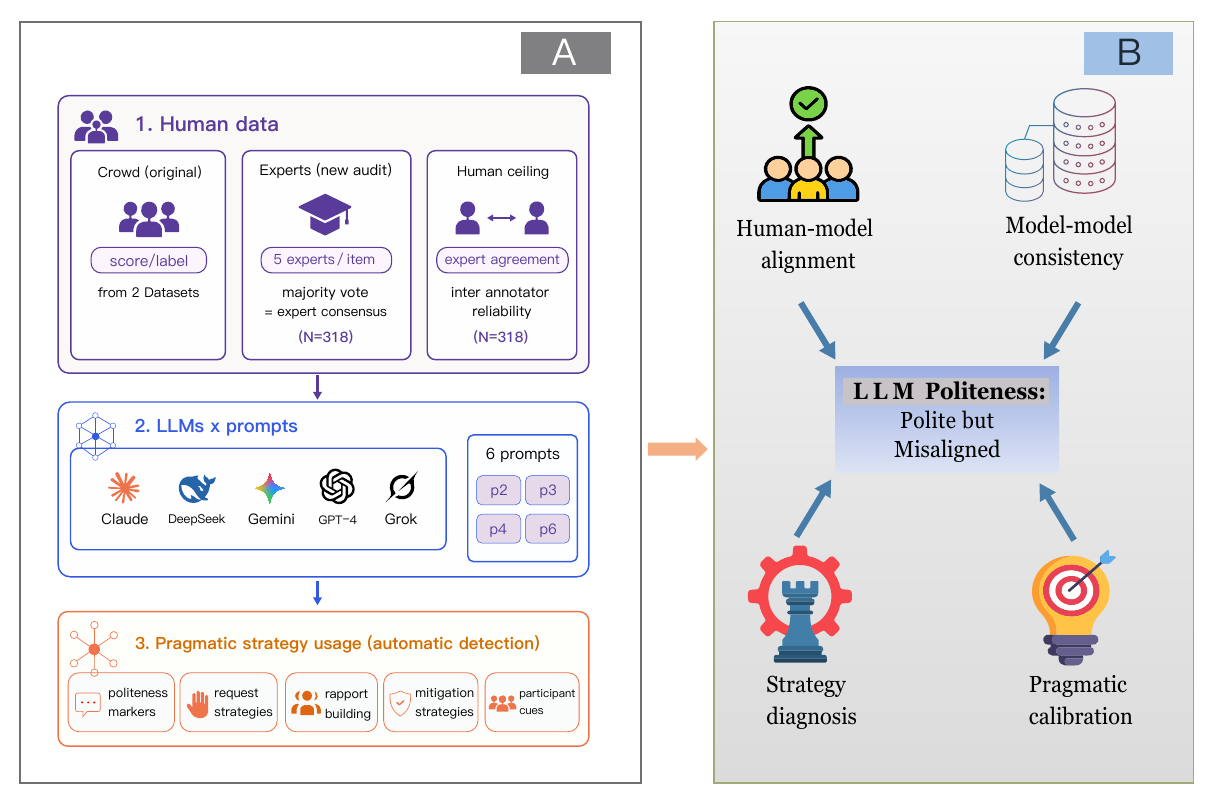}
\caption{Study overview. Panel A shows Dataset~1 continuous scoring with
six prompts, Dataset~2 three-way classification with four prompts
(P2, P3, P4, and P6), and the 318-case expert audit. Panel B shows
model--human alignment, inter-model consistency, strategy differences,
and human-reference comparisons.}
\label{fig:roadmap}
\vspace{-0.2cm}
\end{figure*}

Politeness plays a central role in pragmatics and social interactions, reflecting how speakers manage social relationships, express respect, and mitigate face-threatening acts \cite{leech1983pragmatics,brown1987politeness}. Human judgments of politeness depend not only on lexical markers such as \textit{please} or \textit{thank you}, but also on indirectness, social roles, contextual expectations, face management and cultural norms. As large language models (LLMs) become widely used in conversational agents, customer service systems and educational technologies, it is important to assess whether they evaluate politeness in ways that align with human judgments.

Current LLM benchmarks primarily assess factual knowledge, reasoning, and
safety-related behavior. Benchmarks such as MMLU and BIG-Bench do not directly evaluate pragmatic competence \cite{hendrycks2021measuring,srivastava2023beyond}, while toxicity and
harmfulness benchmarks provide only indirect evidence about politeness,
primarily when impoliteness overlaps with offensive or abusive language
\cite{gehman2020realtoxicityprompts,hartvigsen2022toxigen,
zhang2024safetybench}. Production-based studies show that LLMs can generate conventionally polite language while differing systematically from humans in how they select and deploy politeness strategies \cite{zhao2025comparing}. Even when model outputs conform to conventional politeness cues, such
behavioral conformity does not, by itself, establish that the models'
politeness judgments align with those of humans.

Recent work has expanded the evaluation of pragmatic abilities in LLMs
across a range of phenomena and tasks \cite{sravanthi2024pub,hu2025pragmatics}. Although valuable for assessing
breadth, such evaluations provide limited insight into model behavior
within any single pragmatic phenomenon. They often aggregate heterogeneous phenomena, making it difficult to isolate the mechanisms underlying any single pragmatic failure mode. 

To complement broad pragmatic evaluations with a more fine-grained
diagnosis, we focus on politeness and compare LLM judgments with human
annotations in two English-language online datasets, one with continuous
ratings and the other with three-way categorical labels. Because
politeness judgments are subjective, model--human disagreement may reflect
variation among human evaluators rather than model error or annotation
noise alone \cite{plank2022human,mostafazadeh2022disagreements}. We compare crowd labels, expert consensus, and LLM predictions on
a diagnostic subset of the categorically labeled dataset. This
multi-reference analysis examines whether observed disagreement persists
across human references without treating either reference as a definitive
pragmatic standard.

We address three research questions. First, to what extent do LLM politeness judgments align with
human judgments across models and prompt conditions, and do models
agree more strongly with one another than with humans? Second, which
observable linguistic strategies and systematic patterns
characterize model--human alignment and misalignment? Within this
analysis, we examine how explicit markers and rapport-building strategies are associated with model--human alignment and misalignment. Third, on a diagnostic subset of high-disagreement cases, how does comparison with expert consensus change the interpretation of crowd--LLM disagreement, and do model tendencies toward Neutral judgments persist when expert labels are used as the
reference? Figure~\ref{fig:roadmap} summarizes the datasets, experiments,
and evaluation dimensions.

We contribute (i) a multi-reference evaluation of pragmatic alignment spanning crowd, expert, and LLM politeness judgments; (ii) evidence that LLMs agree more with each other than with humans; (iii) two formalized signatures of pragmatic miscalibration: \textit{neutral compression} and \textit{surface-cue accumulation}; and (iv) a full release of prompts, metrics, model outputs,
and analysis scripts.\footnote{\url{https://github.com/rong4ivy/llm-politeness}}


\section{Related Work}

Our work builds on three lines of research: computational politeness, pragmatic evaluation of LLMs, and human--AI alignment in social contexts.

\subsection{Computational Politeness}

Computational politeness research has progressed from rule-based accounts to data-driven modeling. The Stanford Politeness Corpus \cite{danescu2013computational} operationalized politeness theory \cite{brown1987politeness} through lexical and syntactic cues such as indirection, deference, impersonalization, and modality, showing that politeness can be computationally modeled and linked to social power. Subsequent work extended politeness analysis to supervised and weakly supervised learning, contextualized models, and sociolinguistic applications \cite{priya2024computational}, including dialogue systems \cite{firdaus2020incorporating}, mental health agents \cite{mishra2023help}, cross-cultural communication \cite{kitao1990study}, and feature-based toolkits \cite{yeomans2018politeness}. Whereas prior studies have mainly focused on politeness detection, generation, or feature extraction, we evaluate whether SOTA LLMs produce politeness judgments aligned with human pragmatic norms.

\subsection{Pragmatic Competence in LLMs}

Recent work suggests that LLMs remain limited in pragmatic reasoning. Benchmarks and diagnostic studies show persistent human--model gaps in implicature, presupposition, reference, and deixis \cite{sravanthi2024pub}, as well as difficulties with irony and sarcasm in disagreement \cite{shulginov2025evaluating}, context-dependent impoliteness \cite{andersson2025can}, and prompt-politeness effects \cite{yin2024respect}. Production-based analyses further suggest that LLMs may overuse negative politeness strategies relative to humans \cite{zhao2025comparing}. Complementing these generation- and task-based studies, we directly compare LLM politeness judgments with both continuous human ratings and categorical annotations.

\subsection{Human--AI Alignment in Social Contexts}

Human--AI alignment increasingly concerns social and cultural behavior, not only task performance \cite{ji2024ai}. Although RLHF can reshape model behavior toward human preferences
\cite{ouyang2022training}, recent evaluations report model--human
differences in value priorities, attitudes toward socially important
issues, collective reasoning, and personality-conditioned behavior
\cite{lau2024evaluating,bojic2025towards,qian2024mask,
tsai2024assessing}. Yet pragmatic phenomena such as politeness remain largely absent from standard benchmarks such as MMLU \cite{hendrycks2021measuring} and BIG-Bench \cite{srivastava2023beyond}. We address this gap by testing whether LLM politeness judgments align with human annotations rather than merely exhibiting surface-level polite behavior.

\section{Datasets}

We evaluate LLM politeness judgments using two English-language datasets
with complementary annotation formats: continuous human ratings and
three-way categorical labels.

\noindent\textbf{Dataset 1: Stanford Politeness Corpus.}
The Stanford Politeness Corpus \cite{danescu2013computational} contains
10,957 two-sentence requests: 4,353 from Wikipedia Talk pages and 6,604
from Stack Exchange. Each request was rated by five crowd annotators on
a continuous scale from very impolite to very polite. The corpus
politeness score is the mean of the five ratings after normalization
within annotator, with higher values indicating greater perceived
politeness. We sampled 3,000 requests using the original continuous
human scores to obtain balanced coverage across the observed
politeness-score range. The original continuous scores were retained as
the reference values for the continuous-scoring evaluation.

\noindent\textbf{Dataset 2: Three-way Politeness Corpus.}
For the categorical evaluation, we use the
\texttt{frfede/politeness-corpus} dataset distributed through Hugging
Face \citep{frfede2024politeness}. The repository contains 16,428 text
instances labeled as Impolite, Neutral, or Polite, with 5,476 instances
per category. We sampled 3,000 instances, with 1,000 from each
reference-label category, for the three-way classification evaluation.
Because the repository does not provide detailed documentation of the
original data provenance or label-construction procedure, we treat the
three categories as repository-provided reference labels.

\section{Methods}

\subsection{Evaluated Language Models}

We evaluated five contemporary LLMs via the OpenRouter API \url{https://openrouter.ai} between February 2026 and May 2026: \texttt{google/gemini-2.5-flash}, \texttt{openai/gpt-4.1}, \texttt{anthropic/claude-3.5-sonnet}, \texttt{x-ai/grok-3}, and \texttt{deepseek/deepseek-chat}. For Dataset~2, we additionally evaluated two open-weight models: \texttt{meta-llama/llama-3-8b-instruct} and \texttt{meta-llama/llama-3.1-8b-instruct}. All models were queried using their provider-default sampling parameters (e.g., temperature and top-$p$) without explicitly setting a fixed random seed, reflecting standard out-of-the-box performance. Provider routing via OpenRouter was unconstrained to mirror generic API deployment environments. Failed API requests were retried up to three times, and no responses remained unparseable after retrying. No fine-tuning, model-specific prompt calibration, or manual correction of outputs was performed.

\subsection{Prompt Conditions}

We evaluated prompt sensitivity using six zero-shot prompt templates that
varied along three dimensions:
role framing (neutral evaluator vs. pragmatics expert), cue access
(surface-only vs. pragmatic-function-aware), and output format
(continuous score vs. categorical label).

P1 restricted evaluation to observable linguistic cues. P2 provided a minimal instruction for three-way classification. P3 framed the model as a pragmatics expert and supplied politeness-theoretic criteria. P4 provided an explicit checklist of lexical, syntactic, and discourse features. P5 used a minimal instruction for continuous scoring, whereas P6 elicited scores along five pragmatic dimensions and combined them into an overall score. All prompt templates were held constant across models. The complete templates are provided in Appendix~A.2.


\subsection{Evaluation Metrics}


\noindent\textbf{Dataset~1: Continuous scoring.}
For each model--prompt condition, we measure model--human alignment using
Pearson's correlation coefficient ($r$) and mean absolute error (MAE).
We define the Close Rate as the proportion of instances for which the
absolute difference between the model and human scores is below 0.5.
Pairwise Pearson correlations between model outputs are used to compare
inter-model consistency with model--human alignment.

\noindent\textbf{Dataset~2 (three-class classification).}
We use accuracy, macro-F1, and Cohen's $\kappa$ to measure agreement with the
reference labels, and Fleiss' $\kappa$ to measure agreement across models.
Confusion matrices, label distributions, and the diagnostics below characterize
the direction of disagreement. We compute these metrics separately under P2, P3, P4, and P6 conditions and report both prompt-specific and model-level results.

Overall agreement metrics do not show whether model--reference
disagreements follow a systematic direction. We therefore examine
whether Polite and Impolite labels are shifted toward Neutral, a pattern
we term \textit{neutral compression}. Let $M$ denote the model prediction and
$R$ the reference label, with $N$, $P$, and $I$ denoting Neutral,
Polite, and Impolite, respectively. We define the Neutral Compression
Difference as
\[
\mathrm{NCI}_{diff}
  = \Pr(M=N)-\Pr(R=N),
\]
and the corresponding Neutral Compression Ratio as
\[
\mathrm{NCI}_{ratio}
  = \frac{\Pr(M=N)}{\Pr(R=N)}.
\]
Positive values of $\mathrm{NCI}_{diff}$ and values of
$\mathrm{NCI}_{ratio}>1$ indicate that the model assigns Neutral more
frequently than the reference.

We define the Impoliteness Suppression Difference as
\[
\mathrm{ISR}_{diff}
  = \Pr(R=I)-\Pr(M=I),
\]
where positive values indicate model under-production of Impolite
labels. Finally, we define the class-specific Extreme-to-Neutral Shift
Rate as
\[
\mathrm{ENSR}_{c}
  = \Pr(M=N\mid R=c),
  \qquad c\in\{P,I\}.
\]
Thus, $\mathrm{ENSR}_{P}$ and $\mathrm{ENSR}_{I}$ measure how often
reference-labeled Polite and Impolite instances, respectively, are
shifted to Neutral by the model. These descriptive measures are computed
using either crowd labels or expert consensus as the reference, as
specified in each analysis.


\subsection{Human Baseline Estimation}

To contextualize model--human alignment, we estimated inter-human agreement and a human reference for Dataset~1 using the original annotator-level ratings. Inter-human agreement was computed as the mean pairwise Pearson correlation among the five annotators for each item. We further estimated a leave-one-annotator-out human reference. Each annotator's rating was compared against the mean rating of the remaining four annotators, and the resulting correlations were averaged across annotators. This procedure approximates how well an individual human matches the consensus of other humans and provides a realistic upper bound for model--human comparison on this subjective task. Detailed reliability statistics are reported in Appendix~C.



\subsection{Politeness Strategy Detection}

To examine which observable linguistic strategies are associated with
model--human alignment and misalignment, we applied a rule-based
detector to each input instance. The detector identifies 15 strategies,
including explicit politeness markers, hedges, modal verbs,
interrogatives, conditional constructions, person-reference patterns,
deference, gratitude, greetings, apologies, and positive and negative
lexical cues. Detection is based on lexical lists, regular expressions,
and shallow syntactic patterns. Strategy counts are aggregated at the
instance level and normalized for text length.

For Dataset~1, we define aligned cases as instances with an absolute
model--human score difference below 0.5 and strongly misaligned cases as
instances with a difference of at least 1.0. Instances falling between
these thresholds are excluded from this contrast. For Dataset~2,
aligned and misaligned cases correspond to agreement and disagreement
between model predictions and reference labels, respectively. We
compare strategy frequencies between these groups and examine whether
the direction of each difference is consistent across models and prompt
conditions. Because the analysis involves multiple correlated strategy
comparisons, we treat it as exploratory and emphasize effect direction
and cross-condition consistency rather than isolated significance
tests. Full detection rules and validation results are provided in
Appendix~D.

\subsection{Expert Audit Experiment}

To distinguish model errors from annotation ambiguity, we conducted an
expert audit on 318 Dataset~2 instances sampled from high-disagreement
and strategy-conflict cases. Five proficient L2 English annotators with graduate training in linguistics and pragmatics independently labeled each instance as \textit{Impolite}, \textit{Neutral}, or \textit{Polite} and provided a confidence rating. Majority vote determined the expert consensus label. For each model, a prompt-consensus label was derived as the most frequent
label across the four prompt conditions. Agreement metrics and label
proportions were computed separately for each model using the expert
consensus and crowd labels as references, and were then averaged across
the five models to obtain the mean model-level LLM results reported in
Table~\ref{tab:expert_audit}.
Because the subset was intentionally sampled from difficult cases, it
serves as a diagnostic reference rather than a replacement for the
original crowd annotations.


\section{Results}

\subsection{Result 1: Politeness Scoring}

\begin{figure*}
\centering
\includegraphics[width=0.95\textwidth]{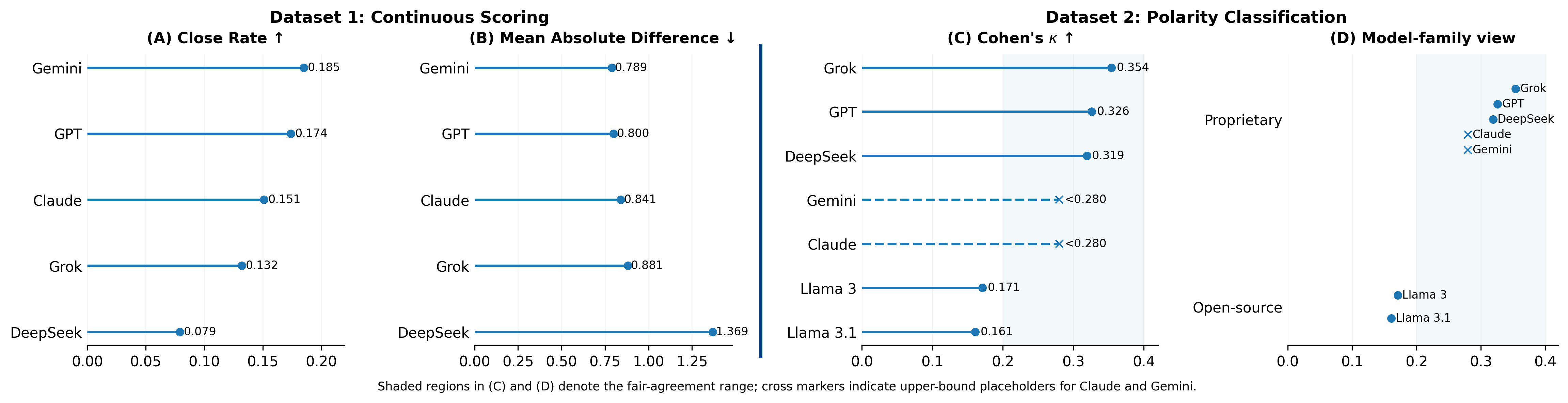}
\caption{Overall LLM--human alignment across continuous scoring and
polarity classification. Panels A--B show Dataset~1 results on the
normalized score scale; Close Rate is the proportion of items with an
absolute model--human difference below 0.5. Panel C reports
model-level Cohen's $\kappa$ values for Dataset~2; colors distinguish
proprietary and open-source models.}
\label{fig:results}
\vspace{-0.1cm}
\end{figure*}

Across all evaluated models, alignment with human politeness scores is limited. The best-performing models achieve only weak-to-moderate correlations with human judgments, and MAE values indicate non-trivial scoring deviations from human consensus. Although many correlations were statistically distinguishable from zero, their magnitudes remained limited. The results are summarized in Panels A \& B of Figure~\ref{fig:results}.

Human agreement provides a useful reference for interpreting model
performance. For the Wikipedia subset of Dataset~1, the mean pairwise
human correlation is $r=0.425$, and Krippendorff's $\alpha=0.424$.
Reliability is higher for the aggregated ratings
(ICC$_{A,k}=0.786$), supporting the use of the consensus score as the
primary human reference. In the leave-one-annotator-out analysis, the
mean human--consensus correlation is $r=0.562$, compared with a best
LLM--human correlation of $r=0.433$. Full human-baseline statistics are
reported in Appendix~\ref{app:human_baseline}.

Across all six prompt conditions, mean pairwise inter-model correlations
exceeded mean model--human correlations. Averaged across prompt
conditions, the mean inter-model correlation was $r=0.724$, compared
with a mean model--human correlation of $r=0.385$. At the prompt level,
mean inter-model correlations ranged from $0.456$ to $0.876$, whereas
mean model--human correlations ranged from $0.308$ to $0.412$, producing
a positive gap of $0.148$ to $0.464$ in every condition. Thus, model
outputs exhibit more similar item-level politeness-score patterns to one
another than to the human reference. This pattern indicates
similar output calibration across the evaluated models. Prompt-level results are reported in Table~\ref{tab:inter_model_correlations} in Appendix~A.3.



Prompt effects were statistically detectable but modest. A mixed-effects
analysis with prompt as a fixed factor and model as a random factor showed a
significant main effect of prompt ($F(5,24)=3.72$, $p=0.012$). Prompt
condition explained 8.3\% of the variance in model--human alignment, compared
with 23.7\% for model identity. Expert prompting produced modest improvements,
but did not close the model--human alignment gap. Full model estimates and
post-hoc comparisons are reported in Appendix~A.4.


Figure~\ref{fig:pattern} summarizes the comparison of politeness strategy usage between aligned and misaligned cases. We found that models align better with humans when sentences employ direct and indicative constructions. In contrast, strategies traditionally associated with politeness, such as gratitude, deference, hedging, modal verbs, first-person framing, and positive affect, are consistently overrepresented in misaligned cases. These effects are stable across models, indicating systematic rather than idiosyncratic biases.

\subsection{Result 2: Politeness Polarity Classification}

We next evaluate LLMs on Dataset~2, where politeness is framed as a three-way classification task. Overall agreement with human annotations is modest. Among proprietary models, Cohen's $\kappa$ ranges from 0.24 to 0.35 despite statistical significance ($p<0.001$): Grok performs best ($\kappa=0.354$), followed by GPT ($\kappa=0.326$) and DeepSeek ($\kappa=0.319$), while Claude and Gemini remain below 0.28. Open-source Llama models perform worse, with Llama~3 ($\kappa=0.171$) and Llama~3.1 ($\kappa=0.161$) falling into the poor-agreement range despite accuracies around 45\%. These results are shown in Panel C of Figure~\ref{fig:results}.


Across the four categorical prompt conditions, agreement among the five
proprietary models ranged from Fleiss' $\kappa=0.518$ to $0.709$.
Mean model-level agreement with the human reference, measured using
Cohen's $\kappa$, ranged from $0.241$ to $0.354$. Although Fleiss' $\kappa$ and Cohen's $\kappa$ summarize different agreement structures, these results provide descriptive evidence that, across all four prompt conditions, model outputs are more mutually
consistent than aligned with the human reference in the categorical
task.

\begin{figure*}
\centering
\includegraphics[width=0.91\textwidth]{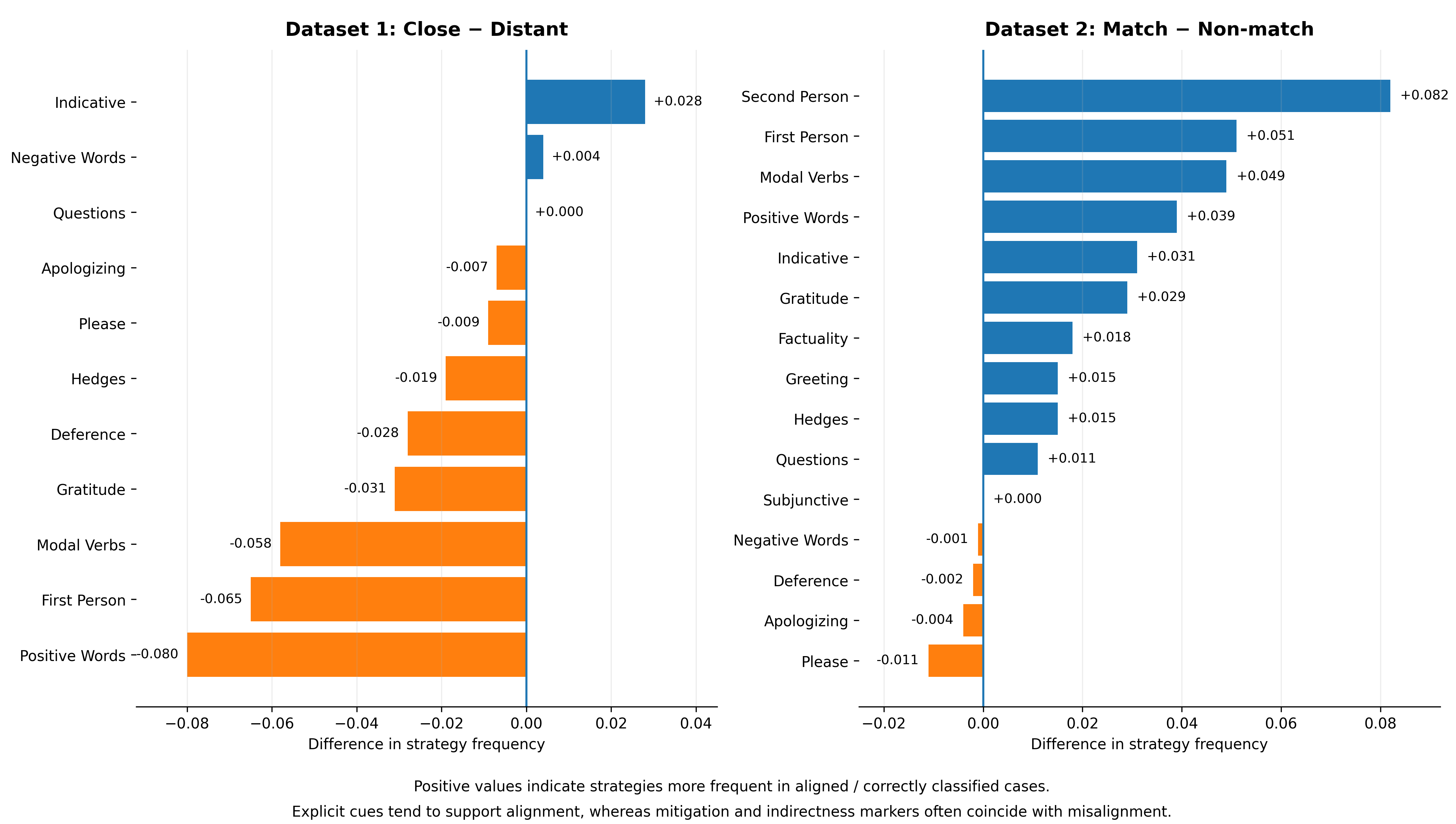}
\caption{Strategy-level differences between aligned and misaligned cases across the two evaluation settings. Positive values indicate strategies more frequent in aligned or correctly classified cases.}
\label{fig:pattern}
\vspace{-0.2cm}
\end{figure*}

Strategy-level analyses further reveal systematic patterns of alignment
and misalignment. As shown in Figure~\ref{fig:pattern}, correctly
classified cases contain more explicit surface cues, including first-
and second-person pronouns, modal verbs, indicative constructions,
positive words, and gratitude expressions. By contrast, indirect or
mitigating strategies such as deference, apologies, and \textit{please}
are slightly more frequent in mismatched cases. Strategy associations differed between Dataset~1 and Dataset~2: some cues
associated with alignment in the categorical evaluation were more frequent
in misaligned cases in the continuous-scoring evaluation.

Confusion-matrix analyses reveal a centralization bias. Human-labeled polite and impolite instances are often classified as neutral, while direct polite--impolite confusions are rare. This pattern is consistent with \textit{neutral compression}, which we use as a descriptive label for the systematic neutralization of extreme politeness judgments rather than as a confirmed causal mechanism. 

To examine whether neutral compression was also present across all evaluated Dataset~2 outputs, we computed NCI and ISR for the full evaluated sample. Across all five proprietary models and four prompt conditions, $\mathrm{NCI}_{diff}$ was positive in every cell (range: $+1.0$ to $+41.2$\,pp; mean $+21.7$\,pp) and $\mathrm{ISR}_{diff}$ was likewise uniformly positive (range: $+7.4$ to $+25.1$\,pp; mean $+16.0$\,pp). This confirms that neutral overproduction and impoliteness suppression are distributional properties of model predictions, not artifacts of the diagnostic subset.

Across both datasets, LLMs align better with humans when politeness is expressed through explicit, low-inference cues, but diverge when judgments require indirectness, social grounding, or pragmatic calibration. Thus, greater lexical politeness does not necessarily improve model--human alignment, indicating a stable surface-cue bias across tasks, models, and prompts.

\subsection{Expert Audit of High-Disagreement Cases}

To further distinguish model errors from annotation ambiguity, we conducted an expert audit on 318 Dataset~2 instances sampled from high-disagreement and strategy-conflict cases. Five linguistically trained experts independently assigned politeness labels and confidence ratings. Expert agreement was fair-to-moderate, with mean pairwise Cohen's $\kappa=0.368$, Fleiss' $\kappa=0.364$, and Krippendorff's $\alpha=0.369$, indicating genuine pragmatic ambiguity in the diagnostic subset.

\begin{table*}[t]
\centering
\small
\caption{Expert-audit results on the diagnostic subset of Dataset~2.
LLM values in Panels A--B are averages across five model-level results,
rather than predictions from a cross-model ensemble. NCI and ISR are
computed from marginal label distributions; ENSR measures directional
extreme-to-Neutral shifts. pp = percentage points.}
\label{tab:expert_audit}
\begin{tabular}{|l|c|c|c|}
\hline
\textbf{Panel A: Agreement comparison} & \textbf{Accuracy} & \textbf{Cohen's $\kappa$} & \textbf{Macro-F1} \\
\hline
Mean model-level LLM vs. Expert consensus & 0.636 & 0.395 & 0.557 \\
\hline
Mean model-level LLM vs. Crowd label & 0.163 & -0.252 & -- \\
\hline
Crowd label vs. Expert consensus & 0.314 & -0.029 & 0.331 \\
\hline
\textbf{Panel B: Label distribution} & \textbf{Neutral} & \textbf{Polite} & \textbf{Impolite} \\
\hline
Crowd labels & 31.4\% & 38.1\% & 30.5\% \\
\hline
Expert consensus & 44.1\% & 31.4\% & 24.6\% \\
\hline
Mean model-level LLM & 68.4\% & 26.6\% & 5.0\% \\
\hline
\textbf{Panel C: Neutral-compression diagnostics} & \textbf{vs. Expert} & \textbf{vs. Crowd} & \textbf{Interpretation} \\
\hline
$\mathrm{NCI}_{diff}$ & +24.3 pp & +37.0 pp & Neutral overproduction \\
\hline
$\mathrm{NCI}_{ratio}$ & 1.55 & 2.18 & Relative Neutral inflation \\
\hline
$\mathrm{ISR}_{diff}$ & +19.6 pp & +25.5 pp & Impolite under-detection \\
\hline
$\mathrm{ENSR}_{imp/pol}$ & 26.9\% / 11.0\% & -- & Extreme-to-Neutral shift \\
\hline
\end{tabular}
\vspace{-0.4cm}
\end{table*}

As shown in Table~\ref{tab:expert_audit}, agreement with expert consensus
was higher than agreement with crowd labels when averaged across the five
model-level comparisons. Mean LLM--expert agreement reached 63.6\%
accuracy, Cohen's $\kappa=0.395$, and macro-F1 of 0.557, whereas mean
LLM--crowd agreement reached 16.3\% accuracy and
$\kappa=-0.252$. Crowd--expert agreement was also limited
(31.4\% accuracy; $\kappa=-0.029$). Because LLM--crowd disagreement was part of the sampling criterion, these
agreement values are diagnostic rather than full-dataset estimates. More generally, these comparisons show that the
estimated degree of model--human agreement depends substantially on the
human reference used.

However, expert alignment does not eliminate systematic model bias. Neutral-compression diagnostics show that LLMs overproduced Neutral labels relative to both experts (+24.3 percentage points; $\mathrm{NCI}_{ratio}=1.55$) and crowd annotators (+37.0 points; $\mathrm{NCI}_{ratio}=2.18$). They also sharply under-detected impoliteness. Only 5.0\% of LLM predictions were Impolite, compared with 24.6\% of expert labels and 30.5\% of crowd labels, yielding $\mathrm{ISR}=+19.6$ and $+25.5$ points, respectively. Directional mismatch analysis further showed stronger neutralization of expert-labeled impolite than polite cases ($\mathrm{ENSR}_{imp}\approx26.9\%$ vs. $\mathrm{ENSR}_{pol}\approx11.0\%$). Thus, the dominant model-specific bias is not simple politeness inflation, but compression of socially marked judgments toward Neutral.

This pattern is important because it separates two forms of disagreement. The higher LLM--expert than LLM--crowd agreement suggests that some model--crowd mismatch reflects annotation ambiguity rather than simple model failure. However, the neutral-compression diagnostics show that LLMs still instantiate a distinct calibration regime. On this diagnostic subset, model predictions show higher agreement with expert consensus than with the crowd labels. Models continue to overproduce Neutral labels and underpredict Impolite
labels.

In sum, these results suggest that crowd annotators, experts, and LLMs instantiate partially distinct pragmatic reference systems. Crowd labels reflect lay social intuitions, expert labels incorporate theory-informed judgments of pragmatic function, and LLM labels follow a stable but lexically anchored calibration pattern. Additionally, the ablation study and error analysis are reported in Appendices~B and~E, while detailed expert-audit results are provided in Appendix~F. These analyses offer further diagnostic evidence for the main findings.

\section{Discussion}
\subsection{Politeness Comprehension in LLMs}

Our results show that current LLMs do not reliably reproduce human politeness judgments. Even the best-performing models achieve only weak correlations with human scores and relatively high MAE values. At the same time, LLMs are more similar to one another than to human judgments. These findings extend broader evaluations showing that
LLM performance remains uneven across pragmatic phenomena
\cite{sravanthi2024pub, hu2025pragmatics}. The high inter-model correlations observed here indicate convergence in item-level scoring
patterns, but such convergence should not be interpreted as evidence of
human-like pragmatic competence.

This model-specific politeness norm is characterized by two systematic biases. 
First, LLMs tend to treat politeness as an additive set of explicit
markers, overweighting surface cues while missing distinctions among
appropriate politeness, overformality, insincerity, and condescension.
This aligns with reported limitations in LLMs' recognition of
context-sensitive impoliteness \cite{andersson2025can}.
Second, LLMs show \textit{neutral compression}, a tendency to collapse socially marked judgments toward Neutral. We do not claim that RLHF causally produces this pattern. Rather, neutral compression is a behavioral signature that may arise from multiple factors, including
post-training objectives, learned label priors, and uncertainty in mapping
utterances to categorical labels.

The expert audit further qualifies the interpretation of model--human
misalignment. On the selected high-disagreement subset, mean model-level
predictions agreed more strongly with expert consensus than with crowd
labels, showing that estimated model--human alignment depends partly on
the human reference. This result does not imply expert-level pragmatic
competence, because the audit subset was intentionally enriched for high-conflict cases, and neutral compression persisted under the expert reference. Models assigned
Neutral labels 1.55 times as often as experts, while their Impolite
prediction rate was approximately one-fifth of the expert rate
(5.0\% vs.\ 24.6\%). Thus, changing the human reference increased measured
agreement but did not eliminate the directional bias toward Neutral.



We use \textit{synthetic pragmatics} to describe this pattern: not a complete absence of pragmatic sensitivity, but a stable model-specific regime revealed by the contrast among crowd, expert, and LLM judgments. Rather than modeling politeness as a context-sensitive social function, LLMs often approximate it through surface-cue accumulation and neutral compression. This interpretation is further supported by ablation and error analyses (Appendices~B and~E), which show reduced sensitivity to pragmatic thresholds, over-reliance on lexical markers, and limited recovery of variance under expert prompting.


\subsection{Linear Pragmatic Accumulation}

The strategy-level analysis suggests a second behavioral signature:
\textit{linear pragmatic accumulation}. LLMs align better with humans when politeness is expressed through explicit surface cues, such as indicative constructions and direct questioning, but diverge on strategies requiring pragmatic inference, including modalization, first-person framing, positive affect, and face management. This pattern suggests that models often approximate
politeness as an additive inventory of lexical and syntactic markers rather than as a context-sensitive social function.

Our findings suggest that politeness judgments cannot be reduced to the
accumulation of explicit cues, and additional politeness markers may make an
utterance appear overformal, insincere, or sarcastic rather than more
polite. In contrast, LLMs appear to treat additional cues as monotonically positive, making them less sensitive to pragmatic boundaries. The expert-audit cases reinforce this interpretation. The hardest errors involve distinguishing appropriate politeness from over-politeness, sincerity from strategic exaggeration, and indirect mitigation from vague or evasive language. Prompt analyses further
show that expert and pragmatic-function-aware prompts can change calibration and inter-model consistency, but do not close the model--human alignment gap.

From the perspectives of Gricean pragmatics, Relevance Theory, and
politic behavior, these failures can be interpreted as reflecting
limited sensitivity to communicative efficiency, contextual
appropriateness, and social baselines
\cite{grice1975logic,sperber1986relevance,locher2005politeness}. One possible contributor is preference-based post-training, including RLHF, which may reward explicit, low-inference signals of helpfulness and harmlessness and thereby make surface politeness markers disproportionately salient
\cite{cheng2025social,rosen2025perils}. We treat this as a plausible
explanation rather than a demonstrated causal mechanism. Future alignment should move beyond scalar rewards toward pragmatic objectives that model graded social appropriateness, penalize excessive politeness, and reward context-sensitive calibration \cite{dahlgren2025helpful,li2025optimizing}.

\subsection{Implications for AI Evaluation and Understanding}

Our findings suggest that politeness alignment should not be treated as agreement with a single human standard. Crowd judgments, expert interpretations, and LLM predictions instantiate partially distinct pragmatic reference systems: lay social intuition, expert pragmatic interpretation, and model-specific lexical calibration. This distinction matters for AI evaluation because polite-sounding LLM outputs do not by themselves establish human-aligned judgment. Polite, safe, or helpful behavior may reflect surface-level behavioral filtering rather than an internalized representation of social norms \cite{ruis2023gold}. Similar issues may extend to other socially grounded
capacities, including fairness, empathy, and respectfulness.

More broadly, politeness illustrates why AI evaluation should distinguish behavioral alignment from socially grounded pragmatic competence. Current benchmarks emphasize knowledge, reasoning, and safety, while under-testing pragmatic abilities such as politeness. Our results therefore caution against equating aligned behavior with sociolinguistic competence.

Finally, our causal interpretation is deliberately limited. We identify neutral compression as a behavioral signature, not as direct evidence of RLHF-induced failure. Safety-oriented post-training is one plausible contributor, but other factors, including pretraining distributions, supervised instruction tuning, label granularity, prompt framing, and pragmatic uncertainty, may also contribute. Because we cannot observe proprietary training pipelines, reward models, system prompts, or safety filters, we treat neutral compression as a robust pattern to be explained rather than a demonstrated causal mechanism.


\section{Conclusion}

We presented a systematic evaluation of LLM politeness judgments against
human annotations. Across the evaluated settings, LLMs exhibit a stable, model-specific politeness norm that is internally consistent and partially expert-like, yet biased toward Neutral compression and explicit surface-cue weighting. Although current LLMs show partial sensitivity to overt politeness cues, their judgments do not consistently align with human annotations in cases requiring context-sensitive pragmatic interpretation. We argue that
politeness should be treated as a first-class evaluation target in future
benchmarks. More broadly, our findings suggest that evaluating trustworthy and socially aligned AI requires moving beyond surface-level output patterns to consider pragmatic function, social relations, and communicative context.


\section*{Limitations}

This study has several limitations. First, both datasets are English-centered and drawn from online interactions, so the findings may not generalize to other languages, cultures, or offline settings where politeness norms differ. Second, the datasets provide limited conversational context, making it difficult to separate model error from missing information about speaker roles, social distance, or interactional history.

Third, the expert audit is diagnostic rather than distributional. The 318 expert-annotated cases were sampled from high-disagreement and strategy-conflict instances, so they should not be used to estimate full-dataset error prevalence. Expert agreement was also only fair-to-moderate, reflecting the inherent subjectivity of pragmatic judgment.

Finally, neutral compression should be interpreted as a behavioral signature, not as a causal claim about RLHF. Although the pattern is consistent with safety-oriented or preference-based post-training, we cannot observe proprietary training data, reward models, system prompts, or safety filters. Other factors, including pretraining distributions, instruction tuning, label granularity, and prompt uncertainty, may also contribute.


 \section*{Ethics Statement}
This study uses publicly accessible datasets containing existing
English-language online text. The expert audit was conducted by five
proficient L2 users of English with graduate-level training in
Linguistics or Applied Linguistics. The annotators evaluated existing
text instances; the study did not collect new personal data from the
original text authors. Because politeness judgments are culturally and
contextually dependent, both the dataset annotations and expert
judgments represent particular linguistic and cultural perspectives.
The findings should therefore not be generalized to other languages,
communities, or cultural settings without further validation.


\bibliography{references}

\appendix

  \begin{center}
    {\LARGE\bfseries Appendix}
  \end{center}
  \vspace{0.8em}
\appendix

\section{Prompt Templates}

\subsection{Overview}

The six prompt framings vary in evaluator role, access to pragmatic
criteria, and output requirements. They serve as sensitivity conditions
for examining whether pragmatic scaffolding affects model--human
agreement, score dispersion, or inter-model consistency. For
Dataset~1, all six framings used a continuous-score output instruction.
For Dataset~2, P2, P3, P4, and P6 used a three-way categorical output
instruction. Because some conditions differ along multiple dimensions,
cross-prompt differences are interpreted descriptively rather than as
causal effects of individual prompt components. The exact task-specific
prompts are included in the accompanying code release.

\subsection{Prompt Framing Templates}

\paragraph{P1: Objective cue-based evaluation.}
Restricts the judgment to observable linguistic cues and explicitly excludes
inferences about speaker intent, emotion, and social context.
\begin{quote}
\small
\textit{Prompt template:}\medskip

\texttt{You are an objective linguistic analyzer. Your task is to evaluate the politeness of text based ONLY on observable linguistic cues such as word choice, sentence structure, and grammatical patterns.}

\texttt{Do NOT infer speaker intent, emotional state, or social context. Do NOT consider who the speaker or addressee might be. Focus exclusively on the linguistic features present in the text itself.}

\texttt{Rate the politeness of the following text on a continuous scale from -3.0 (very impolite) to +2.5 (very polite). Provide only the numerical score with one decimal place.}

\texttt{Text: [INPUT\_TEXT]}

\texttt{Politeness score:}
\end{quote}

\paragraph{P2: Minimal-instruction evaluation.}
Provides no evaluator role, theoretical framework, or feature checklist and
serves as the minimally scaffolded categorical baseline.
\begin{quote}
\small
\textit{Prompt template:}\medskip

\texttt{Evaluate the politeness of the following text and assign one of these labels: Polite, Neutral, or Impolite.}

\texttt{Text: [INPUT\_TEXT]}

\texttt{Politeness score/labels:}
\end{quote}

\paragraph{P3: Expert-pragmatics evaluation.}
Adds an expert role and explicit politeness-theory guidance, thereby testing
whether domain framing changes the elicited judgment.
\begin{quote}
\small
\textit{Prompt template:}\medskip

\texttt{You are a computational sociolinguist and expert in pragmatics, specializing in politeness theory and face-work. Your expertise includes Brown and Levinson's framework of positive and negative politeness strategies, Grice's cooperative principle, and cross-cultural politeness norms.}

\texttt{Evaluate the politeness of the following text using established politeness strategies including but not limited to:}
\begin{itemize}
\item \texttt{Negative politeness: hedging, indirectness, deference, apology, minimizing imposition}
\item \texttt{Positive politeness: solidarity markers, inclusiveness (we/us), compliments, shared identity}
\item \texttt{Face-threat mitigation: modal verbs, subjunctive mood, interrogative framing}
\item \texttt{Conventional markers: please, thank you, greeting formulas}
\end{itemize}

\texttt{Consider the strategic function of linguistic features, not just their surface form. For example, excessive hedging may signal insincerity, and strategic negative framing (``I hate to impose'') may be polite despite containing negative words.}

\texttt{Provide a politeness rating from -3.0 (very impolite) to +2.5 (very polite) based on your expert analysis. Include a brief justification (1--2 sentences) explaining the key factors influencing your rating.}

\texttt{Text: [INPUT\_TEXT]}

\texttt{Rating and Justification:}
\end{quote}

\paragraph{P4: Heuristic-guided evaluation.}
Supplies a checklist of lexical, syntactic, and discourse cues without assigning
an explicit expert identity.
\begin{quote}
\small
\textit{Prompt template:}\medskip

\texttt{Evaluate the politeness of the following text systematically using these linguistic heuristics:}

\textit{Lexical Markers:}
\begin{itemize}
\item \texttt{Explicit politeness markers: please, thank you, sorry, excuse me}
\item \texttt{Positive sentiment: helpful, great, appreciate, kind}
\item \texttt{Negative sentiment: wrong, problem, unfortunately, bad}
\end{itemize}

\textit{Syntactic Patterns:}
\begin{itemize}
\item \texttt{Questions vs.\ direct commands (``Could you...?'' vs.\ ``Do this.'')}
\item \texttt{Modal verbs indicating possibility/permission (can, could, may, might, would)}
\item \texttt{Conditional/subjunctive framing (``If you could...'', ``I was wondering...'')}
\end{itemize}

\textit{Discourse Features:}
\begin{itemize}
\item \texttt{Hedging: perhaps, possibly, maybe, somewhat, I think, it seems}
\item \texttt{Deference: acknowledging expertise/status of addressee}
\item \texttt{Indirectness: avoiding direct imposition or criticism}
\item \texttt{Person reference: use of we/us (inclusive) vs.\ you (direct)}
\end{itemize}

\texttt{Analyze the text for presence of these features, considering both their individual occurrence and their combination. Note that excessive use of multiple strategies may signal insincerity.}

\texttt{Based on this systematic analysis, label the text as: Polite, Neutral, or Impolite.}

\texttt{Text: [INPUT\_TEXT]}

\texttt{Politeness score/labels:}
\end{quote}

\paragraph{P5: Direct continuous-rating command.}
Uses the shortest continuous-scoring instruction and supplies neither an
evaluator role nor analytic criteria.
\begin{quote}
\small
\textit{Prompt template:}\medskip

\texttt{Rate the politeness of the following text on a continuous scale.}

\texttt{Scale: -3.0 (very impolite) to +3.5 (very polite)}

\texttt{Text: [INPUT\_TEXT]}

\texttt{Politeness score}
\end{quote}

The $+3.5$ upper endpoint above is retained from the evaluated P5 prompt and
differs from the $+2.5$ endpoint in P1 and P3. Scores were normalized before
cross-prompt comparison, as described in the main Methods.

\paragraph{P6: Weighted multi-dimensional scoring.}
Decomposes the judgment into five explicit dimensions and a fixed weighted
aggregation rule.
\begin{quote}
\footnotesize
\textit{Prompt template:}\medskip

\texttt{Evaluate the politeness of the following text across five dimensions, each scored from 1 (very low) to 10 (very high):}

\textit{Dimension 1: Strategy Use (Weight: 25\%)}
\begin{itemize}
\setlength{\itemsep}{0pt}\setlength{\parskip}{0pt}\setlength{\parsep}{0pt}\setlength{\topsep}{2pt}
\item \texttt{Presence and appropriateness of politeness strategies: hedging, gratitude, deference, indirectness, positive framing}
\item \texttt{Score 1--3: No strategies or inappropriate use}
\item \texttt{Score 4--6: Minimal or generic strategies}
\item \texttt{Score 7--9: Multiple appropriate strategies}
\item \texttt{Score 10: Sophisticated, contextually calibrated strategy use}
\end{itemize}

\textit{Dimension 2: Contextual Appropriateness (Weight: 20\%)}
\begin{itemize}
\setlength{\itemsep}{0pt}\setlength{\parskip}{0pt}\setlength{\parsep}{0pt}\setlength{\topsep}{2pt}
\item \texttt{Fit between politeness level and implied social context}
\item \texttt{Score 1--3: Inappropriate for any context (too formal or too casual)}
\item \texttt{Score 4--6: Acceptable but generic}
\item \texttt{Score 7--9: Well-calibrated to implied context}
\item \texttt{Score 10: Perfectly contextually tuned}
\end{itemize}

\textit{Dimension 3: Empathy and Consideration (Weight: 20\%)}
\begin{itemize}
\setlength{\itemsep}{0pt}\setlength{\parskip}{0pt}\setlength{\parsep}{0pt}\setlength{\topsep}{2pt}
\item \texttt{Recognition of addressee's face needs, time, and autonomy}
\item \texttt{Score 1--3: Inconsiderate or demanding}
\item \texttt{Score 4--6: Neutral, minimal consideration}
\item \texttt{Score 7--9: Thoughtful acknowledgment of imposition}
\item \texttt{Score 10: Exceptional empathy and perspective-taking}
\end{itemize}

\textit{Dimension 4: Indirectness vs.\ Clarity (Weight: 20\%)}
\begin{itemize}
\setlength{\itemsep}{0pt}\setlength{\parskip}{0pt}\setlength{\parsep}{0pt}\setlength{\topsep}{2pt}
\item \texttt{Balance between face-saving indirectness and communicative efficiency}
\item \texttt{Score 1--3: Too direct (rude) or too indirect (confusing)}
\item \texttt{Score 4--6: Moderate directness, lacks optimization}
\item \texttt{Score 7--9: Well-balanced indirectness}
\item \texttt{Score 10: Optimal pragmatic clarity and politeness}
\end{itemize}

\textit{Dimension 5: Register and Tone (Weight: 15\%)}
\begin{itemize}
\setlength{\itemsep}{0pt}\setlength{\parskip}{0pt}\setlength{\parsep}{0pt}\setlength{\topsep}{2pt}
\item \texttt{Appropriateness of formality level and emotional tone}
\item \texttt{Score 1--3: Inappropriate register (too casual/formal)}
\item \texttt{Score 4--6: Generic neutral register}
\item \texttt{Score 7--9: Context-appropriate formality}
\item \texttt{Score 10: Perfectly calibrated register and tone}
\end{itemize}

\texttt{Compute the weighted average of the five dimension scores:}

\texttt{Overall Score = (D1 $\times$ 0.25) + (D2 $\times$ 0.20) + (D3 $\times$ 0.20) + (D4 $\times$ 0.20) + (D5 $\times$ 0.15)}

\texttt{Then convert the 1--10 scale to the politeness scale using this mapping:}
\begin{itemize}
\setlength{\itemsep}{0pt}\setlength{\parskip}{0pt}\setlength{\parsep}{0pt}\setlength{\topsep}{2pt}
\item \texttt{1.0--2.5 $\rightarrow$ -3.0 to -2.0 (very impolite)}
\item \texttt{2.6--4.5 $\rightarrow$ -1.9 to -0.5 (impolite)}
\item \texttt{4.6--6.5 $\rightarrow$ -0.4 to +0.4 (neutral)}
\item \texttt{6.6--8.5 $\rightarrow$ +0.5 to +1.5 (polite)}
\item \texttt{8.6--10.0 $\rightarrow$ +1.6 to +2.0 (very polite)}
\end{itemize}

\texttt{Text: [INPUT\_TEXT]}

\texttt{Provide \texttt{Politeness score/labels:}: (1) Five dimension scores, (2) Weighted average, (3) Final politeness score}
\end{quote}

\subsection{Cross-Prompt Analysis}

To assess prompt sensitivity, we computed pairwise correlations between model
scores under all six prompts for each model and then averaged the correlations
across models. Scores were first placed on the common normalized scale used in
the main analysis. Table~\ref{tab:prompt_correlation} therefore summarizes
similarity between observed outputs under different prompt formulations.

\begin{center}
\centering
\captionof{table}{Inter-prompt correlation matrix averaged across the five tested
models. Higher values indicate more similar score patterns across prompt
formulations; they do not establish prompt invariance.}
\label{tab:prompt_correlation}
\small
\resizebox{\columnwidth}{!}{%
\begin{tabular}{|l|c|c|c|c|c|c|}
\hline
& \textbf{P1} & \textbf{P2} & \textbf{P3} & \textbf{P4} & \textbf{P5} & \textbf{P6} \\
\hline
\textbf{P1} & 1.00 & 0.87 & 0.72 & 0.84 & 0.94 & 0.79 \\
\hline
\textbf{P2} & 0.87 & 1.00 & 0.69 & 0.91 & 0.86 & 0.75 \\
\hline
\textbf{P3} & 0.72 & 0.69 & 1.00 & 0.68 & 0.74 & 0.81 \\
\hline
\textbf{P4} & 0.84 & 0.91 & 0.68 & 1.00 & 0.83 & 0.73 \\
\hline
\textbf{P5} & 0.94 & 0.86 & 0.74 & 0.83 & 1.00 & 0.80 \\
\hline
\textbf{P6} & 0.79 & 0.75 & 0.81 & 0.73 & 0.80 & 1.00 \\
\hline
\textbf{Avg} & 0.83 & 0.82 & 0.73 & 0.80 & 0.83 & 0.78 \\
\hline
\end{tabular}
}
\end{center}

The average inter-prompt correlation is 0.80, indicating that the broad ranking
of items is fairly stable across prompt formulations. The minimally scaffolded
or atheoretical conditions P1, P2, and P5 are especially similar (mean pairwise
$r=0.89$), with P1 and P5 showing the strongest pairwise correlation
($r=0.94$). P3 has the lowest mean correlation with the other conditions
($r=0.73$), suggesting that expert framing changes the output pattern more than
the other tested formulations. P6 occupies an intermediate position, with
moderate correlations to both minimally framed and more structured prompts.
These results support the use of multiple prompts as a robustness check: the
outputs are neither prompt-invariant nor reorganized completely by the tested
instructions. They do not, however, identify the internal reasoning process
responsible for the differences.

\paragraph{Inter-model versus model--human correlations.}
For each prompt condition, we computed the ten pairwise Pearson
correlations among the five models and the five corresponding
model--human correlations using the same evaluated instances.
Table~\ref{tab:inter_model_correlations} reports the mean correlations
and their differences by prompt. Inter-model correlations are higher
under every prompt condition, although the magnitude of the gap varies
across prompts.

\begin{table}[H]
\centering
\small
\caption{Mean model--human and pairwise inter-model Pearson correlations
for Dataset~1 by prompt condition. Gap is the inter-model mean minus the
model--human mean.}
\label{tab:inter_model_correlations}
\begin{tabular}{lccc}
\hline
\textbf{Prompt} &
\textbf{Model--human $r$} &
\textbf{Inter-model $r$} &
\textbf{Gap} \\
\hline
P1 & 0.398 & 0.729 & +0.331 \\
P2 & 0.388 & 0.741 & +0.353 \\
P3 & 0.393 & 0.706 & +0.313 \\
P4 & 0.308 & 0.456 & +0.148 \\
P5 & 0.411 & 0.837 & +0.426 \\
P6 & 0.412 & 0.876 & +0.464 \\
\hline
Overall & 0.385 & 0.724 & +0.339 \\
\hline
\end{tabular}
\end{table}

\subsection{Statistical Testing of Prompt Effects}

We tested prompt effects on model--human correlation using prompt condition as
a fixed factor and model identity as a random factor. The omnibus effect was
statistically detectable, $F(5,24)=3.72$, $p=.012$, but modest in magnitude:
the reported variance decomposition attributes 8.3\% of variation in alignment
to prompt condition, compared with 23.7\% to model identity and 68.0\% to
residual variation. Tukey-adjusted comparisons locate the clearest differences
in the more explicitly scaffolded conditions. P3 exceeds P1 by
$\Delta r=0.045$ ($p=.008$) and P2 by $\Delta r=0.042$ ($p=.011$), while P6
exceeds P2 by $\Delta r=0.038$ ($p=.019$). No tested difference among P1, P2,
P4, and P5 reaches significance (all $p>.10$).

The statistical and correlational analyses therefore lead to the same
high-level conclusion: prompt formulation changes elicited judgments, but the
changes are smaller than the differences associated with model identity and do
not close the model--human alignment gap. P3 and P6 are useful sensitivity
conditions because they expose the models to explicit pragmatic structure;
P2 and P5 remain necessary baselines for determining how much of that behavior
appears without such scaffolding. The separate ablation in Appendix~B further
examines score dispersion under expert framing, but is not treated as an
isolated causal test of any single prompt component.

\section{Prompt-Framing Ablation}
\label{app:prompt_ablation}

This follow-up ablation examines whether prompt framing changes the dispersion
and human alignment of continuous politeness scores. It is separate from the
six-prompt sensitivity analysis in Appendix~A. We use a stratified subset of
1,000 Dataset~1 requests spanning impolite ($<-0.5$), neutral
($[-0.5,0.5]$), and polite ($>0.5$) human-rated cases. Three system-prompt
conditions are compared: a default assistant prompt (Baseline), an expert
pragmatics prompt (Expert), and a neutral linguistic-analysis prompt (Neutral).
All conditions use the same scoring instruction and a scale from $-3.0$ to
$+2.5$; scores are min--max normalized to $[-1,+1]$ before comparison. The
same transformation is applied to all conditions, so normalization is a
preprocessing step rather than an ablated factor.

We evaluate GPT-4.1, Claude-3.5-Sonnet, and Gemini-2.5-Flash using four
complementary measures: score variance, interquartile range (IQR), Pearson
correlation with human ratings, and a score-centralization ratio. The latter is
the proportion of model predictions in $[-0.3,0.3]$ divided by the
corresponding human proportion, so values above 1.0 indicate that model scores
are more concentrated near the center than the human ratings. We call this
continuous-score pattern \textit{score centralization} to distinguish it from
the categorical neutral-compression measures used for Dataset~2.

\begin{table}[!ht]
\centering
\caption{Prompt-framing ablation on the Dataset~1 subset. Centralization is
the ratio of model to human predictions in the normalized interval
$[-0.3,0.3]$; values above 1.0 indicate greater model concentration near the
center. Pearson $r$ measures agreement with human ratings.}
\label{tab:ablation_summary}

\footnotesize
\setlength{\tabcolsep}{3pt}
\resizebox{\columnwidth}{!}{%
\begin{tabular}{|l|l|c|c|c|c|}
\hline
\textbf{Model} &
\textbf{Condition} &
$\boldsymbol{\sigma^2}$ &
\textbf{IQR} &
\textbf{Central.} &
\textbf{$r$} \\
\hline

\multirow{3}{*}{GPT-4.1}
& Baseline & 0.42 & 0.85 & 1.73 & 0.433 \\
\cline{2-6}
& Expert & 0.51 & 1.02 & 1.42 & 0.478 \\
\cline{2-6}
& Neutral & 0.47 & 0.93 & 1.58 & 0.451 \\
\hline

\multirow{3}{*}{Claude-3.5}
& Baseline & 0.38 & 0.78 & 1.81 & 0.392 \\
\cline{2-6}
& Expert & 0.46 & 0.95 & 1.48 & 0.429 \\
\cline{2-6}
& Neutral & 0.41 & 0.84 & 1.67 & 0.407 \\
\hline

\multirow{3}{*}{Gemini-2.5}
& Baseline & 0.45 & 0.89 & 1.65 & 0.417 \\
\cline{2-6}
& Expert & 0.52 & 1.05 & 1.38 & 0.456 \\
\cline{2-6}
& Neutral & 0.48 & 0.95 & 1.53 & 0.433 \\
\hline

\multirow{3}{*}{\textbf{Average}}
& Baseline & 0.42 & 0.84 & 1.73 & 0.414 \\
\cline{2-6}
& Expert & \textbf{0.50} & \textbf{1.01}
         & \textbf{1.43} & \textbf{0.454} \\
\cline{2-6}
& Neutral & 0.45 & 0.91 & 1.59 & 0.430 \\
\hline
\end{tabular}%
}
\end{table}
Table~\ref{tab:ablation_summary} shows the same ordering across all three
models. Relative to the baseline condition, expert framing increases score
variance by approximately 19\% and IQR by 19.9\% on average, reduces score centralization
by 17.5\%, and increases Pearson correlation with human ratings by 0.040. The
Neutral condition changes the same measures in the same direction but less
strongly. Thus, the expert condition produces more dispersed scores and
slightly greater human correlation, but the expert-condition centralization
ratios remain above 1.0 (1.38--1.48) and the correlations remain moderate.

The category-conditioned GPT-4.1 results clarify where the additional
dispersion occurs. Under expert framing, the mean score for human-rated
Impolite cases changes from $-0.42$ (SD $=0.31$) to $-0.68$ (0.39), while the
mean for Polite cases changes from $+0.71$ (0.35) to $+0.95$ (0.41). Neutral
cases change only from $+0.08$ (0.28) to $+0.05$ (0.32). The separation
between the Polite and Impolite means therefore increases from 1.13 to 1.63
points (44.2\%). Because this breakdown concerns GPT-4.1 only, it is treated as
an illustration of the aggregate pattern rather than a cross-model result.

\subsection{Strategy-Level Effects}

We also examined how the change under expert prompting varied across selected
strategy categories. As shown in Table~\ref{tab:ablation_strategies}, the largest changes
occurred for multiple hedges, first-person plural framing, strategic negative
wording, casual register, and subtle deference. In contrast, \textit{please},
gratitude, overt negative words, and indicative forms changed little.

\begin{table}[h]
\centering
\caption{Effect of expert prompting on selected politeness strategies.
$\Delta r$ is the change in Pearson correlation with human ratings.}
\label{tab:ablation_strategies}
\small
\resizebox{\columnwidth}{!}{%
\begin{tabular}{|l|c|c|c|}
\hline
\textbf{Strategy} & \textbf{Baseline $r$} & \textbf{Expert $r$} & $\boldsymbol{\Delta r}$ \\
\hline
\multicolumn{4}{|l|}{\textit{Larger increase in $r$}} \\
\hline
Multiple hedges & 0.31 & 0.42 & +0.11 \\
\hline
First-person plural & 0.28 & 0.38 & +0.10 \\
\hline
Strategic negative words & 0.24 & 0.33 & +0.09 \\
\hline
Casual register & 0.35 & 0.43 & +0.08 \\
\hline
Subtle deference & 0.37 & 0.44 & +0.07 \\
\hline
\multicolumn{4}{|l|}{\textit{Minimal change}} \\
\hline
Simple gratitude & 0.52 & 0.54 & +0.02 \\
\hline
Please & 0.49 & 0.50 & +0.01 \\
\hline
Overt negative words & 0.58 & 0.59 & +0.01 \\
\hline
Indicative forms & 0.61 & 0.61 & 0.00 \\
\hline
\end{tabular}
}
\end{table}

The changes are therefore not uniform across the detected strategies. The
largest correlation increases occur in categories whose interpretation may
depend more heavily on pragmatic function, although this analysis does not
identify the process producing those changes.

\subsection{Interpretation}

The ablation supports two conclusions about the evaluated continuous-scoring
setting. First, expert framing changes the distribution of model scores: it
increases dispersion, reduces score centralization, and is associated with
slightly higher model--human correlations. Second, these changes are limited.
Even under expert framing, model scores remain more centralized than the human
ratings and only moderately correlated with them. The experiment therefore
does not distinguish an elicitation ceiling from a representational or
training-related limitation.

This analysis does not directly test the categorical NCI, ISR, or ENSR measures
used to define neutral compression in Dataset~2. It instead shows that the
related but distinct concentration of continuous scores near the center is
partly sensitive to prompt framing. The result does not identify a training
cause: safety-oriented post-training, pretraining distributions, instruction
tuning, label granularity, and task uncertainty remain possible explanations.

\section{Human Reference Analysis}
\label{app:human_baseline}

Annotator-level ratings are available for the Wikipedia subset of Dataset~1.
We use them for two complementary reference analyses. Inter-annotator
statistics measure agreement among the five individual ratings, whereas the
leave-one-annotator-out analysis compares each annotator with the mean of the
other four. Table~\ref{tab:human_reference_summary} reports both analyses in a
single summary.

\begin{center}
\centering
\footnotesize
\captionof{table}{Human reference analysis on the Wikipedia subset of Dataset~1.
Panel A reports inter-annotator agreement. Panel B reports leave-one-annotator-out
agreement, averaged across held-out annotators. MAE and RMSE are shown on the
original 1--25 scale and on the approximate normalized scale used in
Figure~\ref{fig:results}.}
\label{tab:human_reference_summary}
\textbf{Panel A: Inter-annotator agreement}\\[2pt]
\begin{tabular}{lc}
\hline
\textbf{Metric} & \textbf{Value} \\
\hline
Pairwise Pearson $r$ & $0.425 \pm 0.013$ \\
Krippendorff's $\alpha$ & 0.424 \\
ICC$_{A,1}$ (single rater) & 0.424 \\
ICC$_{A,k}$ (average rater) & 0.786 \\
Within-utterance variance & 57.6\% \\
Between-utterance variance & 42.4\% \\
\hline
\end{tabular}
\\[5pt]
\textbf{Panel B: Leave-one-annotator-out agreement}\\[2pt]
\begin{tabular}{lc}
\hline
\textbf{Metric} & \textbf{Value} \\
\hline
Pearson $r$ & 0.562 \\
Spearman $r$ & 0.538 \\
MAE / RMSE (original) & 3.118 / 4.030 \\
MAE / RMSE (normalized) & $\approx 0.669$ / $\approx 0.865$ \\
3-class accuracy & 55.7\% \\
Macro-F1 & 0.542 \\
Cohen's $\kappa$ & 0.310 \\
\hline
\end{tabular}
\end{center}

\paragraph{Interpretation.}
The single-rater measures indicate only moderate agreement: pairwise
correlation, Krippendorff's $\alpha$, and single-rater ICC are all approximately
0.42. Reliability increases to ICC$_{A,k}=0.786$ when the five ratings are
aggregated, supporting the use of the corpus mean as the primary reference
while also showing why a single annotator should not be treated as definitive.
The variance decomposition points to the same limitation: within-utterance
variation (57.6\%) exceeds between-utterance variation (42.4\%) on this subset.

In the leave-one-annotator-out analysis, an individual annotator correlates
with the remaining-annotator consensus at $r=0.562$ on average. This is higher
than the mean pairwise correlation because the four-person reference averages
over some annotator-specific variation. The corresponding three-class
agreement remains moderate (55.7\% accuracy; macro-F1 $=0.542$;
$\kappa=0.310$), illustrating that discretization does not remove disagreement
in the underlying judgments. These values contextualize the model results, but
they are not a performance ceiling: the held-out annotator and the aggregated
corpus score are different estimands, and both remain sensitive to annotator
composition and scale construction.

\section{Politeness Strategy Detection System}

To characterize observable linguistic cues associated with LLM--human
alignment and misalignment, we implemented a rule-based politeness strategy
detector. Its cue inventory draws on Brown and Levinson's politeness theory
\cite{brown1987politeness} and related work on hedging, modality, and sentiment
\cite{fraser2010pragmatic,palmer2001mood,liu2012survey}. We use explicit rules
rather than a learned classifier so that the operational definitions can be
inspected and applied consistently across datasets. The detector is intended
as a diagnostic instrument, not as a complete model of pragmatic function.

The detector identifies 15 strategies grouped into five broad categories:
(i) explicit politeness markers, including \textit{please}, gratitude,
apologies, and greetings; (ii) syntactic mitigation strategies, including
hedges, modal verbs, interrogatives, and subjunctive or conditional framing;
(iii) person-reference strategies, including first-person, second-person, and
deference markers; (iv) semantic polarity, including positive and negative
words; and (v) directness and factuality, including indicative constructions
and information-focused language. The strategy names are shorthand for the
surface patterns targeted by the rules. They neither constitute an exhaustive
taxonomy of politeness nor imply that the source datasets were annotated using
Brown and Levinson's framework.

Detection is based on regular expressions, curated keyword lists, shallow
syntactic patterns, and sentiment lexicons. For each request, we tokenize and
segment the text, apply the strategy rules at the sentence level, aggregate
detections to the request level, and normalize strategy counts by text length.
We then compare length-normalized strategy frequencies between aligned and
misaligned cases.
For Dataset~1, aligned cases are those with model--human score differences
below 0.5, while strongly misaligned cases have differences of at least 1.0.
Cases between these thresholds are excluded from this contrast.
For Dataset~2, aligned and misaligned cases correspond to label agreement and
disagreement, respectively. Figure~\ref{fig:pattern} reports the difference in
mean strategy frequency between the aligned and misaligned groups, so positive
values indicate that a detected cue is more frequent in aligned cases.

To validate the detector, we manually coded the presence or absence of each
strategy in 200 randomly sampled requests from Dataset~1 and compared these
labels with the detector outputs. As shown in
Table~\ref{tab:detection_validation}, the macro averages across the 15
strategies are Cohen's $\kappa=0.87$, precision $=0.89$, and recall $=0.88$.
Agreement is highest for explicit markers such as \textit{please}, gratitude,
and questions, and lower for deference and factuality.

\begin{table}[h]
\centering
\caption{Validation of the rule-based strategy detector against human
annotations on 200 requests. Precision and recall treat human annotations as
the reference.}
\label{tab:detection_validation}
\small
\begin{tabular}{|l|c|c|c|}
\hline
\textbf{Strategy} & $\boldsymbol{\kappa}$ & \textbf{Precision} & \textbf{Recall} \\
\hline
Please & 0.97 & 0.98 & 0.99 \\
\hline
Gratitude & 0.94 & 0.96 & 0.95 \\
\hline
Apologizing & 0.91 & 0.93 & 0.92 \\
\hline
Greeting & 0.89 & 0.91 & 0.90 \\
\hline
Hedges & 0.82 & 0.85 & 0.84 \\
\hline
Modal verbs & 0.88 & 0.90 & 0.89 \\
\hline
Questions & 0.95 & 0.97 & 0.96 \\
\hline
Subjunctive & 0.79 & 0.81 & 0.82 \\
\hline
First person & 0.93 & 0.95 & 0.94 \\
\hline
Second person & 0.94 & 0.96 & 0.95 \\
\hline
Deference & 0.76 & 0.79 & 0.78 \\
\hline
Positive words & 0.86 & 0.88 & 0.87 \\
\hline
Negative words & 0.84 & 0.86 & 0.85 \\
\hline
Indicative & 0.81 & 0.83 & 0.84 \\
\hline
Factuality & 0.74 & 0.77 & 0.76 \\
\hline
\textbf{Macro average} & \textbf{0.87} & \textbf{0.89} & \textbf{0.88} \\
\hline
\end{tabular}
\end{table}

The validation results show that the rules recover the manually coded cue
categories with high average agreement; they do not validate a particular
pragmatic interpretation of every detected occurrence. The detector may still
miss implicit strategies, context-dependent realizations, or cases in which
the absence of a marker is pragmatically meaningful. The associations in the
main analysis are also task-dependent. For example, gratitude and modal verbs
are more frequent in misaligned cases in the continuous-scoring analysis but
more frequent in correctly classified cases in the categorical analysis.
The resulting evidence complements the qualitative error analysis in
Appendix~E and the prompt sensitivity analyses in Appendices~A and~B.

\section{Error Analysis}

To complement the aggregate results, we qualitatively examined selected
high-agreement and high-disagreement cases from the continuous-scoring analysis
of Dataset~1. The examples illustrate contrasts involving explicit markers,
rapport-building cues, strategic negative wording, and possible perceptions of
excess or insincerity. They were selected diagnostically and therefore do not
estimate the prevalence of these patterns. Moreover, the original ratings do
not provide annotator rationales; the descriptions below are post-hoc
interpretations of the observed score differences.

\subsection{Successful Alignment}

Table~\ref{tab:high_agreement} shows selected cases in which model scores
closely match human judgments and the text contains explicit signals of
politeness or impoliteness.

\begin{table*}[h]
\centering
\caption{Selected examples of close model--human agreement in the Dataset~1
continuous-scoring analysis.}
\label{tab:high_agreement}
\small
\begin{tabular}{|p{0.9\linewidth}|}
\hline
\textbf{Human \& Model: Polite} \\
\hline
``Could you please help me understand this issue? I would really appreciate your guidance.'' \\
Human: $+1.8$, Model: $+1.7$ \\
Strategies: modal, please, gratitude, indirectness \\
\hline
``Thank you so much for taking the time to look at this. Your expertise would be incredibly valuable here.'' \\
Human: $+1.9$, Model: $+1.8$ \\
Strategies: gratitude, deference, positive lexicon \\
\hline
\textbf{Human \& Model: Impolite} \\
\hline
``You need to fix this immediately. Why are you wasting everyone's time?'' \\
Human: $-1.7$, Model: $-1.6$ \\
Strategies: direct command, second-person address, negative lexicon \\
\hline
``This is completely wrong. Did you even bother to read the documentation?'' \\
Human: $-1.5$, Model: $-1.4$ \\
Strategies: negative lexicon, confrontational question \\
\hline
\end{tabular}
\end{table*}

In these examples, polite requests combine
conventional markers such as \textit{please}, gratitude, modality, and
deference, while impolite examples contain overtly confrontational wording.
The examples show that close agreement can occur when the limited input contains
salient lexical and syntactic evidence.

\subsection{Illustrative Disagreements}

Table~\ref{tab:misalignment} presents selected disagreements involving the
relationship between surface form and possible pragmatic function.

\begin{table*}[h]
\centering
\caption{Selected examples of model--human disagreement in the Dataset~1
continuous-scoring analysis. Interpretations are diagnostic rather than
annotator-provided explanations.}
\label{tab:misalignment}
\small
\begin{tabular}{|p{0.9\linewidth}|}
\hline
\textbf{Type 1: Model overestimates politeness} \\
\hline
``I was wondering if perhaps you might possibly consider taking a look at this when you have a moment?'' \\
Human: $+0.3$, Model: $+1.5$, $\Delta=+1.2$ \\
Possible cue conflict: multiple hedges coincide with a higher model score \\
\hline
``Your work is absolutely amazing and perfect in every way. Could you possibly help me with just this tiny little thing?'' \\
Human: $-0.2$, Model: $+1.3$, $\Delta=+1.5$ \\
Possible cue conflict: exaggerated praise coincides with a higher model score \\
\hline
\textbf{Type 2: Model underestimates politeness} \\
\hline
``Quick question---any thoughts on the best approach here?'' \\
Human: $+1.2$, Model: $+0.1$, $\Delta=-1.1$ \\
Possible cue conflict: casual brevity receives a lower model score \\
\hline
``I see your point. Let me think about how we can address this together.'' \\
Human: $+1.4$, Model: $+0.2$, $\Delta=-1.2$ \\
Possible cue conflict: inclusive framing receives a lower model score \\
\hline
\textbf{Type 3: Lexical-polarity conflicts} \\
\hline
``I would hate to impose, but if it's not too much trouble, might you consider this?'' \\
Human: $+1.6$, Model: $-0.3$, $\Delta=-1.9$ \\
Possible cue conflict: negative wording co-occurs with a mitigating construction \\
\hline
``Just checking in to see if you got my last message.'' \\
Human: $+0.8$, Model: $-0.5$, $\Delta=-1.3$ \\
Possible cue conflict: a brief follow-up receives a lower model score \\
\hline
\end{tabular}
\end{table*}

In these selected cases, multiple hedges and praise terms coincide with higher
model than human scores, whereas casual brevity and inclusive framing coincide
with lower model scores. Strategic negative phrases such as \textit{hate to
impose} illustrate a further tension between lexical polarity and the human
reference. These examples are compatible with several explanations, including
cue accumulation, missing context, and annotation uncertainty; they do not
distinguish among them.

\subsection{Strategy-Conditioned Error Patterns}

We next condition on the presence of each reported strategy category and report the
average absolute difference between human and model scores. This analysis asks
a different question from Figure~\ref{fig:pattern}: the figure compares cue
frequencies between aligned and misaligned groups, whereas
Table~\ref{tab:strategy_misalignment} compares error magnitude among cases
containing a cue. A cue may therefore have a relatively small conditional error
while still occurring more often in the misaligned group.

\begin{center}
\centering
\captionof{table}{Strategy-conditioned error patterns in the Dataset~1
continuous-scoring analysis. $|\Delta|$ is the average absolute model--human
score difference. Consistency is the number of evaluated model--prompt
configurations showing the listed pattern or signed direction. Model
$\uparrow$/$\downarrow$ indicates that model scores are higher/lower than the
human ratings; dashes indicate no dominant signed difference.}
\label{tab:strategy_misalignment}
\small
\resizebox{\columnwidth}{!}{%
\begin{tabular}{|l|c|c|c|}
\hline
\textbf{Strategy} & \textbf{Avg. $|\Delta|$} & \textbf{Signed pattern} & \textbf{Consistency} \\
\hline
\multicolumn{4}{|l|}{\textit{Better aligned strategies}} \\
\hline
Indicative forms & 0.28 & --- & 18/20 \\
\hline
Overt negative lexicon & 0.31 & --- & 19/20 \\
\hline
Simple gratitude & 0.35 & --- & 17/20 \\
\hline
\multicolumn{4}{|l|}{\textit{Poorly aligned strategies}} \\
\hline
Multiple hedges & 0.89 & Model $\uparrow$ & 19/20 \\
\hline
Excessive deference & 0.95 & Model $\uparrow$ & 18/20 \\
\hline
First-person plural & 0.78 & Model $\downarrow$ & 16/20 \\
\hline
Excessive positive words & 1.02 & Model $\uparrow$ & 20/20 \\
\hline
Counterfactual + hedge & 0.84 & Model $\uparrow$ & 17/20 \\
\hline
Casual register & 0.73 & Model $\downarrow$ & 15/20 \\
\hline
Strategic negative words & 1.15 & Model $\downarrow$ & 19/20 \\
\hline
\end{tabular}
}
\end{center}

Within this diagnostic analysis, multiple hedges, excessive positive wording,
and deference coincide with higher model than human scores, while first-person
plural framing, casual register, and strategic negative wording coincide with
lower model scores. 

The signed patterns recur across the 20 evaluated model--prompt
configurations: multiple hedging has a positive model--human difference in 19,
strategic negative wording a negative difference in 19, first-person plural
framing a negative difference in 16, and excessive deference a positive
difference in 18. This consistency describes the evaluated outputs without
identifying a shared internal mechanism.

\paragraph{Connection to the main results.}
The examples and strategy-conditioned results help interpret the task-dependent
associations reported in the main paper: disagreements of similar magnitude can
involve different observable cue configurations, including multiple markers,
rapport-building language, indirect mitigation, and lexical polarity. Because
the examples were selected diagnostically and the strategy categories are
rule-based, this analysis neither estimates the prevalence of these patterns nor
identifies the mechanism that produces them.

\section{Expert Audit Results}
\label{app:expert_audit_results}

Where Appendix~E examines continuous-score errors in Dataset~1, this appendix
reports the detailed expert-audit results summarized in the main text for
Dataset~2. The audit contains 318 high-disagreement and strategy-conflict cases from
Dataset~2, each independently labeled by five linguistically trained experts.
The expert-consensus label is compared with the original crowd label and with
predictions from five models under four prompt conditions. Because the cases
were deliberately selected for diagnostic difficulty, the results do not
estimate error prevalence in the full dataset. Mean LLM agreement scores are
unweighted averages of the five model-level results in
Table~\ref{tab:expert_audit_details}; mean label proportions are likewise
averaged across model-level output distributions and do not represent a single
ensemble prediction.

\begin{figure}[t]
\centering
\includegraphics[width=\columnwidth]{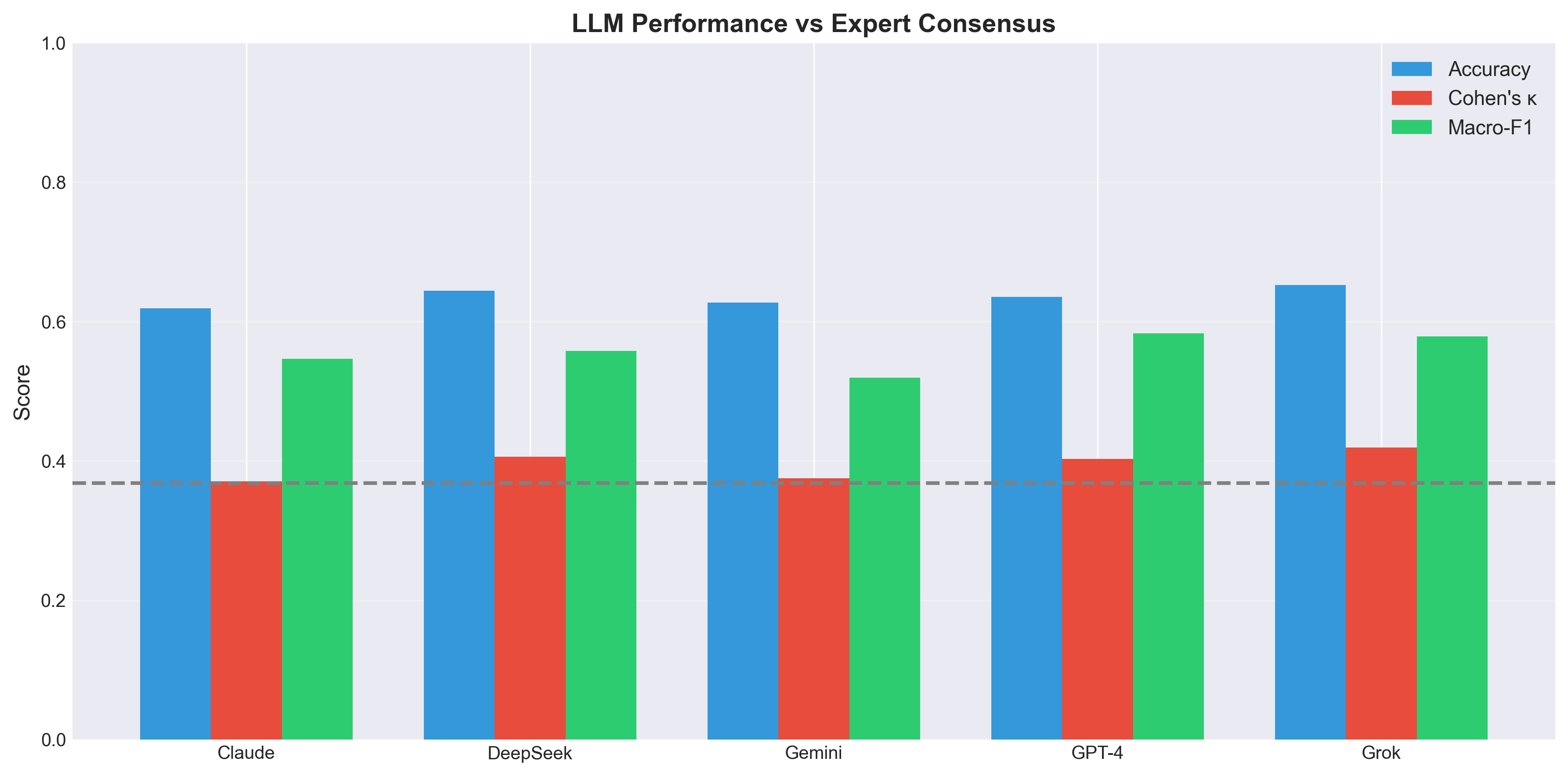}
\caption{Model-level agreement with expert consensus. The dashed line marks
the mean pairwise expert agreement value ($\kappa=0.368$) as a descriptive
reference; it is not a performance ceiling.}
\label{fig:llm_expert_performance}
\end{figure}

\begin{figure}[t]
\centering
\includegraphics[width=\columnwidth]{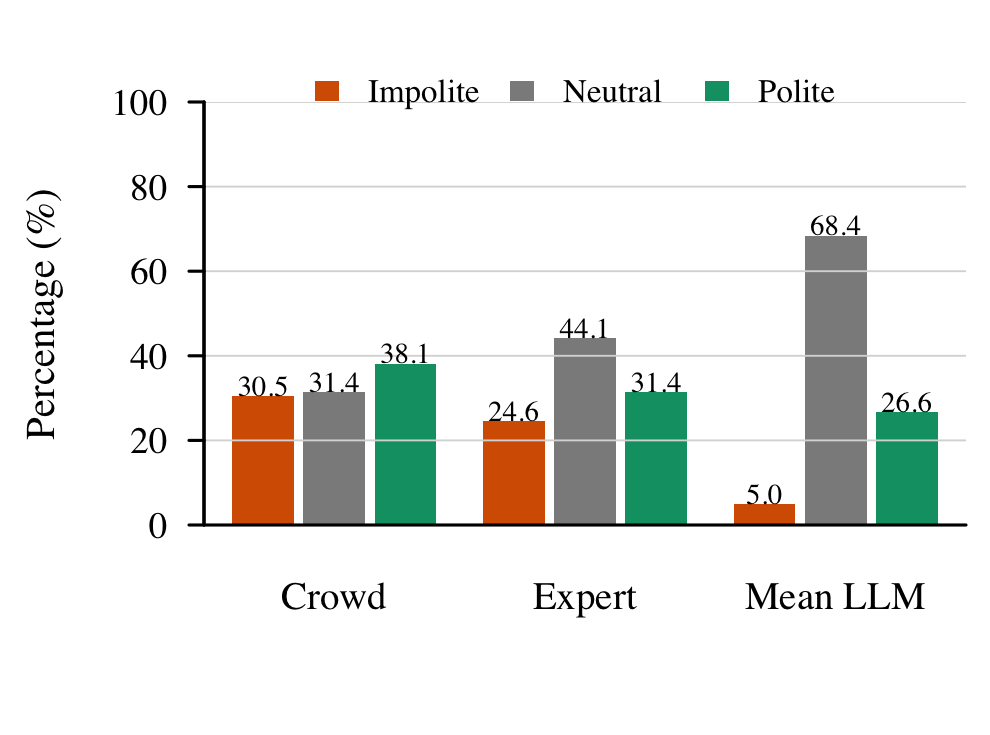}
\caption{Label distributions for crowd annotations, expert consensus, and the
mean LLM distribution on the diagnostic expert-audit subset. Model outputs
are more strongly concentrated in Neutral and contain fewer Impolite
predictions than both human references.}
\label{fig:expert_audit_label_distribution}
\end{figure}

\paragraph{Expert agreement and label distributions.}
Agreement among the experts was limited: mean pairwise Cohen's $\kappa=0.368$,
Fleiss' $\kappa=0.364$, and Krippendorff's $\alpha=0.369$. In total, 53.4\% of
instances met the study's strong- or full-consensus criterion. These values
indicate substantial variation on the deliberately difficult cases, but do not
identify whether that variation arises from item ambiguity, individual
differences, or the annotation protocol. As Figure~\ref{fig:expert_audit_label_distribution}
shows, expert consensus contains more Neutral labels than the crowd reference
(44.1\% vs. 31.4\%), while the mean LLM distribution is more concentrated in
Neutral still (68.4\%). Impolite labels show the reverse ordering: 30.5\% for
the crowd reference, 24.6\% for expert consensus, and 5.0\% for the mean LLM
distribution.

\paragraph{Sensitivity to the human reference.}
Agreement estimates change substantially with the reference used on this
subset. The mean model-level LLM--expert comparison yields 63.6\% accuracy,
$\kappa=0.395$, and macro-F1 of 0.557. The corresponding LLM--crowd comparison
yields 16.3\% accuracy and $\kappa=-0.252$, while crowd--expert agreement is
31.4\% accuracy, $\kappa=-0.029$, and macro-F1 of 0.331. These aggregate values
are already tabulated in Table~\ref{tab:expert_audit} in the main text and are
not repeated here. In particular, the low LLM--crowd agreement is expected
partly by design because high LLM--crowd disagreement was a sampling criterion.
Conversely, the higher LLM--expert agreement neither establishes expert
consensus as ground truth nor demonstrates general expert-level pragmatic
competence.

\paragraph{Model-level agreement.}
Table~\ref{tab:expert_audit_details} (Panel A) shows modest descriptive
variation across models. Grok has the highest agreement with expert consensus
($\kappa=0.419$), followed by DeepSeek ($\kappa=0.406$) and GPT-4.1
($\kappa=0.403$). Prompt consistency is higher than model--expert agreement for
all five models ($\kappa=0.701$--$0.833$), indicating relatively stable outputs
across the four evaluated prompt conditions. Figure~\ref{fig:llm_expert_performance}
places the model-level values alongside mean pairwise expert agreement as a
descriptive reference, not a performance ceiling.

\paragraph{Directional mismatches.}
For this analysis, the majority-vote LLM label is compared with expert
consensus. Among their 118 mismatches, Expert Impolite $\rightarrow$ LLM
Neutral is the most frequent of the three reported directions (21 cases),
followed by Expert Polite $\rightarrow$ LLM Neutral (11) and Expert Neutral
$\rightarrow$ LLM Polite (9; Table~\ref{tab:expert_audit_details}, Panel B).
These counts describe label shifts rather than their linguistic or causal
source. Their percentages use all 118 mismatches as the denominator, whereas
$\mathrm{ENSR}_{I}$ and $\mathrm{ENSR}_{P}$ divide by all expert-labeled
Impolite and Polite cases, respectively.

\begin{table}[H]
\centering
\footnotesize
\caption{Detailed expert-audit results. Panel A reports agreement with expert
consensus by model; prompt consistency is average agreement across four prompt
conditions. Panel B reports the three most frequent majority-LLM/expert-consensus
mismatch directions; percentages use the 118 mismatches, rather than all 318
audited instances, as the denominator.}
\label{tab:expert_audit_details}
\textbf{Panel A: Model-level agreement}\\[2pt]
\resizebox{\columnwidth}{!}{%
\begin{tabular}{lcccc}
\hline
\textbf{Model} & \textbf{Acc.} & \textbf{$\kappa$} & \textbf{Macro-F1} & \textbf{Prompt $\kappa$} \\
\hline
Claude   & 0.619 & 0.371 & 0.546 & 0.701 \\
DeepSeek & 0.644 & 0.406 & 0.558 & 0.728 \\
Gemini   & 0.627 & 0.375 & 0.519 & 0.783 \\
GPT-4.1    & 0.636 & 0.403 & 0.583 & 0.702 \\
Grok     & 0.653 & 0.419 & 0.579 & 0.833 \\
\hline
Average  & 0.636 & 0.395 & 0.557 & 0.749 \\
\hline
\end{tabular}
}\\[5pt]
\textbf{Panel B: Directional mismatches}\\[2pt]
\begin{tabular}{lcc}
\hline
\textbf{Direction} & \textbf{Count} & \textbf{Percent} \\
\hline
Expert I $\rightarrow$ LLM N & 21 & 17.8\% \\
Expert P $\rightarrow$ LLM N & 11 & 9.3\% \\
Expert N $\rightarrow$ LLM P & 9 & 7.6\% \\
\hline
\end{tabular}\\[3pt]
I = Impolite; N = Neutral; P = Polite.
\end{table}

Taken together, the expert audit qualifies rather than resolves model--human disagreement. Changing the human reference changes the estimated agreement, while
the concentration of LLM predictions in Neutral remains visible relative to
both crowd labels and expert consensus. The audit therefore supplies an
alternative diagnostic reference without adjudicating which annotation source
is correct.

\end{document}